%% file: GSPR.tex
\documentclass[letterpaper, 10 pt, conference]{ieeeconf}
\IEEEoverridecommandlockouts    
\usepackage{hyperref}
\usepackage{soul}
\usepackage{makecell}

\usepackage{threeparttable}

\usepackage{caption}
\hypersetup{
    colorlinks=true,
    linkcolor=magenta,
    filecolor=magenta,      
    urlcolor=magenta,
}

\input{stachnisslab-latex}


\usepackage{bm}
\usepackage{multirow}
\usepackage{rotating}  

\title{\LARGE \bf GSPR: Multimodal Place Recognition Using 3D Gaussian Splatting \\ for Autonomous Driving}

\author{Zhangshuo Qi$^{1\dag}$ \and Junyi Ma$^{2\dag}$ \and Jingyi Xu$^2$ \and Zijie Zhou$^1$ \and Luqi Cheng$^1$ \and Guangming Xiong$^{1*}$ 
  \thanks{
  This work was supported by the National Natural Science Foundation of China under Grant 52372404.
  }
  \thanks{
  $^{1}$Zhangshuo Qi, Zijie Zhou, Luqi Cheng and Guangming Xiong are with Beijing Institute of Technology, Beijing, 100081, China
  }
  \thanks{
  $^{2}$Junyi Ma and Jingyi Xu are with Shanghai Jiao Tong University, Shanghai, 200240, China
  }
  \thanks{
  $^{*}$Corresponding author (xiongguangming@bit.edu.cn)
  }
  \thanks{$^\dag$ Equal Contribution}
}

\begin{document}
\maketitle
\pagestyle{empty}  
\thispagestyle{empty} 

\IEEEpeerreviewmaketitle
\thispagestyle{empty}
\pagestyle{empty}

\begin{abstract}

Place recognition is a crucial component that enables autonomous vehicles to obtain localization results in GPS-denied environments. In recent years, multimodal place recognition methods have gained increasing attention. They overcome the weaknesses of unimodal sensor systems by leveraging complementary information from different modalities. However, most existing methods explore cross-modality correlations through feature-level or descriptor-level fusion, suffering from a lack of interpretability. Conversely, the recently proposed 3D Gaussian Splatting provides a new perspective on multimodal fusion by harmonizing different modalities into an explicit scene representation. In this paper, we propose a 3D Gaussian Splatting-based multimodal place recognition network dubbed GSPR. It explicitly combines multi-view RGB images and LiDAR point clouds into a spatio-temporally unified scene representation with the proposed Multimodal Gaussian Splatting. A network composed of 3D graph convolution and transformer is designed to extract spatio-temporal features and global descriptors from the Gaussian scenes for place recognition. Extensive evaluations on three datasets demonstrate that our method can effectively leverage complementary strengths of both multi-view cameras and LiDAR, achieving SOTA place recognition performance while maintaining solid generalization ability. Our open-source code will be released at \url{https://github.com/QiZS-BIT/GSPR}.\\

\end{abstract}

\section{Introduction}
\label{sec:intro}

Given an observation from sensors at the current moment (query), place recognition needs to determine which location in the global map (database) the observation corresponds to. Place recognition is an important module in most navigation systems, capable of correcting accumulated drift in SLAM algorithms and often serving as the first step in global localization. In autonomous driving systems, cameras are commonly used for vision-based place recognition (VPR), providing rich semantics and texture information~\cite{arandjelovic2013all, arandjelovic2016netvlad, keetha2023anyloc}. However, vision features extracted from camera images exhibit lower stability and result in suboptimal recognition accuracy when facing variations in lighting, seasons, and weather in large-scale outdoor environments. In contrast, LiDAR sensors show high stability against these factors, leading to more robust LiDAR-based place recognition (LPR)~\cite{uy2018pointnetvlad, liu2019lpd, ma2022overlaptransformer}. However, the recognition performance of LPR is still limited by the natural sparsity of LiDAR point clouds, and the lack of texture and semantic information.

In recent years, multimodal place recognition (MPR) methods such as MinkLoc++~\cite{komorowski2021minkloc++} and LCPR~\cite{zhou2023lcpr} have demonstrated the potential advantages of fusing data from complementary camera and LiDAR modalities, attracting more research interests. Some MPR methods extract features for each modality independently, followed by descriptor-level fusion. Others generate multimodal descriptors through modality-wise feature-level fusion. However, these methods suffer from a lack of interpretability, which limits insight into cross-modal interactions.
Recently, the introduction of 3D Gaussian Splatting (3D-GS)~\cite{kerbl20233d} provides a new perspective on multimodal fusion. It is proposed to construct an explicit scene representation using 3D Gaussians, effectively capturing the geometry information of the scene. By aggregating temporally continuous observations from multiple views, 3D-GS comprehensively constructs spatial structure representations, providing the possibility of explicable spatio-temporal fusion of multimodal place recognition. 

\begin{figure}
  \centering
  \includegraphics[width=1\linewidth]{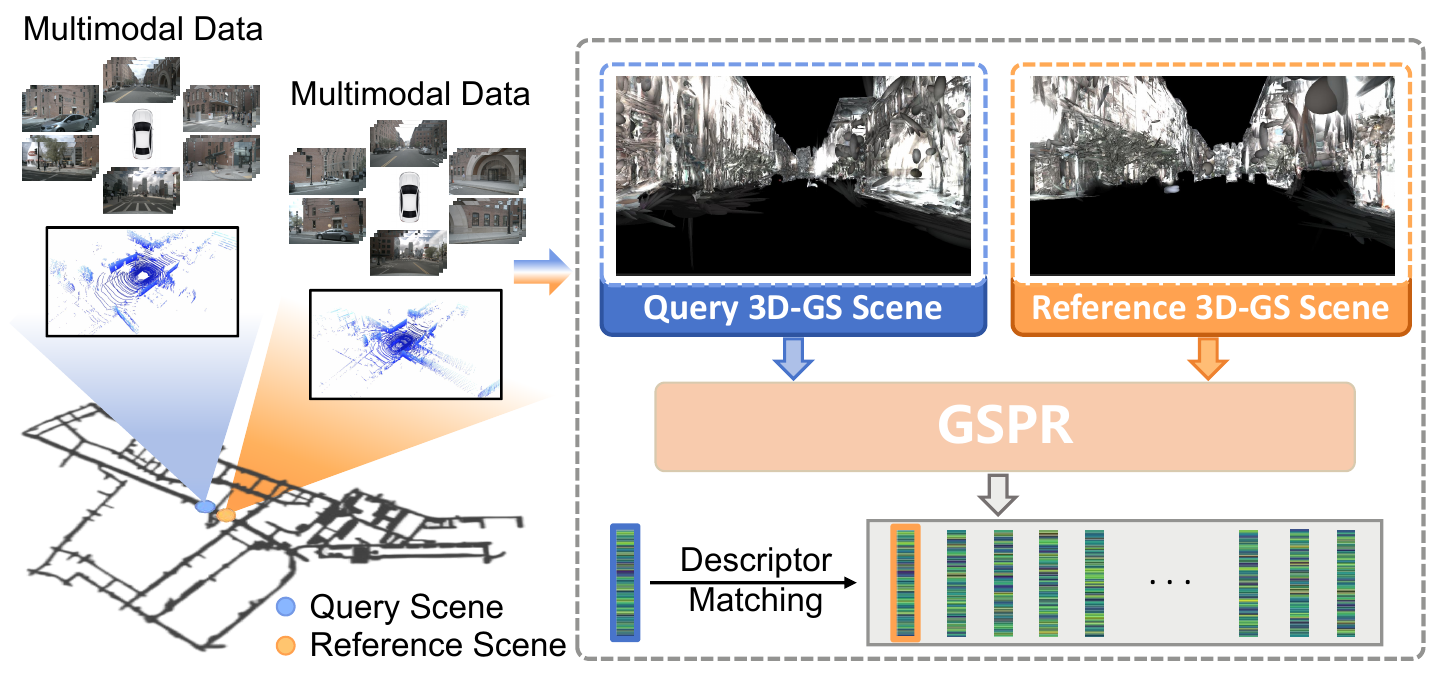}
  \caption{Effectively integrating different modalities is crucial for leveraging multimodal data. GSPR harmonizes multi-view RGB images and LiDAR point clouds into a unified scene representation based on Multimodal Gaussian Splatting. 3D graph convolution and transformer are utilized to extract both local and global spatio-temporal information embedded in the scene, ultimately generating discriminative descriptors.}
  \label{fig:motivation}
  \vspace{-0.6cm}
\end{figure}

In this paper, we propose a 3D Gaussian Splatting-based multimodal place recognition method namely GSPR, as shown in~\figref{fig:motivation}. We first design a Multimodal Gaussian Splatting (MGS) method to represent autonomous driving scenarios. We utilize LiDAR point clouds as a prior for the initialization of Gaussians, which helps to address the failures of structure-from-motion (SfM) in such environments. In addition, a mixed masking mechanism is employed to remove unstable features less valuable for place recognition. 
By taking advantage of the attribute updates and adaptive density control strategies in the Gaussian Optimization process, we can obtain Gaussian scenes that complement the advantages of each modality. In the explicit scene representations, the Gaussians are densely and uniformly distributed, reflecting the fine-grained geometric structure of the scene. Additionally, the Gaussians encode rich semantic and texture information from the images. We then downsample the unordered Gaussians through voxel partitioning, and develop a network based on 3D graph convolution and transformer to extract high-level spatio-temporal features for generating discriminative descriptors for place recognition. Through the proposed MGS, we fuse multimodal data into a unified explicit scene representation, providing the basis for multimodal place recognition.

In summary, our main contributions are as follows:
\begin{itemize}
\item We propose Multimodal Gaussian Splatting method to harmonize multi-view camera and LiDAR data into explicit scene representations suitable for place recognition.
\item We propose GSPR, a novel MPR network equipped with 3D graph convolution and transformer to aggregate local and global spatio-temporal information inherent in the MGS scene representation.
\item Extensive experimental results on three datasets demonstrate that our method outperforms the state-of-the-art unimodal and multimodal methods on place recognition performance while showing a solid generalization ability on unseen driving scenarios.
\end{itemize}

\section{Related Work}
\label{sec:related}

\subsection{Scene Representation in Place Recognition}
\label{sec:scene}
Place recognition is a classic topic in the fields of robotics and computer vision, and there have been various types of traditional methods based on handcrafted descriptors~\cite{milford2012seqslam, arandjelovic2013all, 9944848}. With the rapid development of deep learning, an increasing number of learning-based approaches~\cite{arandjelovic2016netvlad, uy2018pointnetvlad, komorowski2021minkloc++, ma2022overlaptransformer} have been proposed and overall present better recognition performance than traditional counterparts. 

In place recognition tasks, autonomous vehicles perceive the environment through cameras or LiDAR sensors and attempt to build a reasonable scene representation corresponding to the place where the vehicle is situated. The input form of place recognition methods is closely related to the type of sensors. Most vision-based place recognition methods~\cite{arandjelovic2016netvlad, garg2020delta, mereu2022learning, izquierdo2024optimal} treat RGB images as trivial scene representations. NetVLAD~\cite{arandjelovic2016netvlad} aggregates features from RGB images into global descriptors. To enhance robustness to appearance changes, Delta Descriptors~\cite{garg2020delta} constructs change-based descriptors using sequential images. JIST~\cite{berton2023jist} leverages a large uncurated set of images to mitigate the issue of limited sequential data, achieving robust place recognition. LiDAR or Radar-based place recognition~\cite{hui2021pyramid, cait2022autoplace, ma2022seqot, ma2023cvtnet} represents the scene as a point cloud or its various derived forms. For instance, PointNetVLAD~\cite{uy2018pointnetvlad} uses submaps obtained by stacking LiDAR point clouds as the scene representation, Autoplace~\cite{cait2022autoplace} uses BEV views constructed from multi-view radars to capture structural information of the scene, OverlapNet~\cite{chen2022overlapnet} obtains dense depth information by projecting unordered LiDAR point clouds into range images, BVMatch~\cite{luo2021bvmatch} and BEVPlace~\cite{luo2023bevplace} achieve efficient place recognition and pose estimation through LiDAR BEVs, and CVTNet~\cite{ma2023cvtnet} combines multi-layer BEVs and range images to alleviate the information loss of 3D point cloud projection. These primitive scene representations from different modalities each have their own advantages and disadvantages. Our method, using 3D-GS, generates unified explicit scene representations that harmonize different modalities, allowing the representation of both the fine-grained geometric structure and the texture information.

\subsection{Multimodal Place Recognition}
\label{sec:mpr}
Recently, multimodal place recognition has aroused great interest due to its ability to leverage the complementary advantages of multiple sensors. MinkLoc++~\cite{komorowski2021minkloc++} concatenates point cloud descriptors from sparse convolutions with image descriptors from pre-trained ResNet Blocks. AdaFusion~\cite{lai2022adafusion} adjusts the weights of different modalities in the global descriptor through a weight generation branch. LCPR~\cite{zhou2023lcpr} employs multi-scale attention to explore inner feature correlations between different modalities during feature extraction. EINet~\cite{xu2024explicit} introduces a novel multimodal fusion strategy that supervises image feature extraction with LiDAR depth maps and enhances LiDAR point clouds with image texture information. It is notable that most MPR methods fuse abstracted descriptor vectors or integrate features from different modalities to harmonize multimodal data. However, the process by which the two modalities complement and integrate remains neither explicit nor explainable. In our proposed GSPR, we instead explicitly fuse spatio-temporal information from different modalities by Multimodal Gaussian Splatting. The distributions and properties of the optimized Gaussians can reflect the rationality of harmonizing multimodal data, allowing for explicit and thorough exploitation of the spatio-temporal correlations between different modalities.

\begin{figure*}[ht]
  \centering
  \includegraphics[width=1\linewidth]{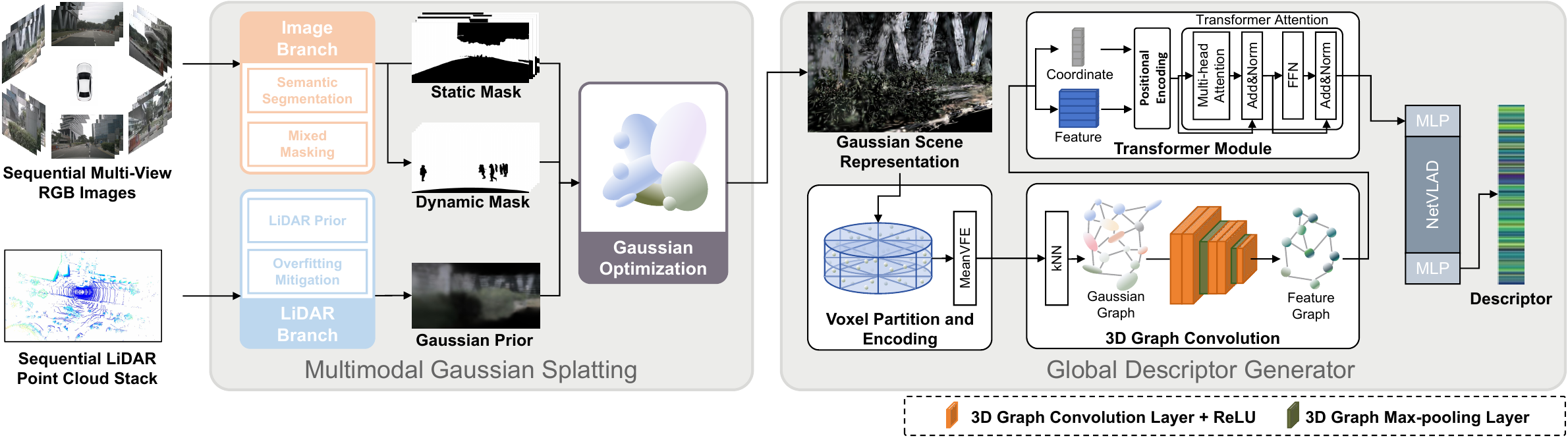}
  \caption{The overall architecture of GSPR. Multimodal Gaussian Splatting employs strategies including LiDAR-based Gaussian initialization and mixed masking mechanism to fuse LiDAR and camera data into a spatio-temporal unified MGS scene representation. The Global Descriptor Generator voxelizes the MGS scene representation and employs 3D graph convolution and transformer to extract high-level local and global spatio-temporal features embedded within the scene. Finally, the high-level spatio-temporal features are aggregated into place recognition descriptors using NetVLAD-MLPs combos.}
  \label{fig:overall}
  \vspace{-0.6cm}
\end{figure*}

\subsection{3D Gaussian Splatting for Autonomous Driving}
\label{sec:gs_driving}
3D-GS performs well in static, bounded small scenes, but faces limitations such as scale uncertainty and training view overfitting in autonomous driving scenarios. To address these challenges, Street Gaussian~\cite{yan2024street} uses LiDAR point prior and introduces 4D spherical harmonics to represent dynamic objects. Driving Gaussian~\cite{zhou2024drivinggaussian} integrates an incremental static Gaussian model with a composite dynamic Gaussian graph for scene reconstruction. Following this, S3Gaussian~\cite{huang2024textit} attempts to eliminate the reliance on annotated data by introducing a multi-resolution hex plane for self-supervised foreground-background decomposition. DHGS~\cite{shi2024dhgs} uses a signed distance field to supervise the geometric attributes of road surfaces. Inspired by these works, we propose Multimodal Gaussian Splatting, leveraging multimodal data and the proposed mixed masking mechanism, to provide stable and geometrically accurate reconstruction results of autonomous driving scenes suitable for place recognition.

\section{Our Approach}
\label{sec:method}

The overview of our proposed GSPR is depicted in~\figref{fig:overall}. GSPR is composed of two components: Multimodal Gaussian Splatting (MGS) and Global Descriptor Generator (GDG). Multimodal Gaussian Splatting fuses the multi-view camera and LiDAR data into a spatio-temporally unified Gaussian scene representation. Global Descriptor Generator extracts high-level spatio-temporal features from the scene through 3D graph convolution and transformer module, and aggregates the features into discriminative global descriptors for place recognition.

\subsection{Multimodal Gaussian Splatting}
\label{sec:multimodal_3dgs}
\begin{figure}[t]
  \centering
  \includegraphics[width=1\linewidth]{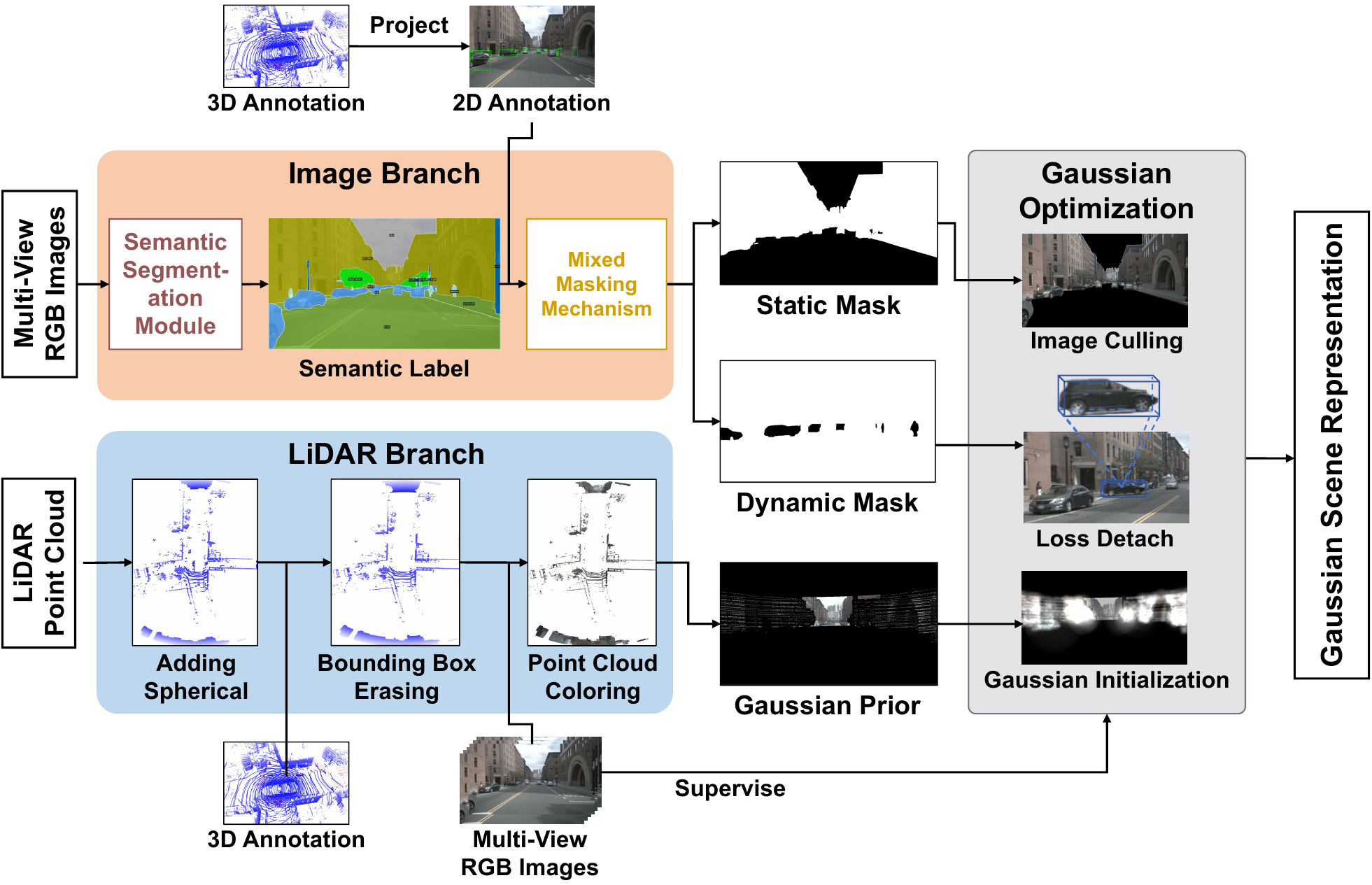}
  \caption{The Multimodal Gaussian Splatting (MGS) pipeline initializes the Gaussians using processed LiDAR point clouds as prior information. RGB image sequences generate masks to guide Gaussian optimization through semantic segmentation and mixed masking. After iterative optimization, the multimodal data are integrated into a unified MGS scene representation.}
  \label{fig:mask}
  \vspace{-0.6cm}
\end{figure}

As illustrated in~\figref{fig:mask}, we introduce Multimodal Gaussian Splatting for autonomous driving scene reconstruction. The method processes multimodal data through the Image Branch and the LiDAR Branch, and then integrates different modalities into a spatio-temporally unified explicit scene representation through Gaussian Optimization. This provides a scene representation with a larger area of coverage and a more uniform distribution than the LiDAR point cloud. Additionally, each Gaussian encodes the features and texture information corresponding to the splatting region in the image, ultimately enabling explicit spatio-temporal fusion of multimodal data.

\textbf{LiDAR prior.} The vanilla 3D-GS uses SfM to reconstruct point clouds for initializing the Gaussian model. However, in autonomous driving scenarios, SfM can fail due to the complexity of the scene, illumination changes, and the high-speed movement of the ego vehicle. To address this, we introduce LiDAR point clouds for initializing the position of Gaussians following~\cite{yan2024street, zhou2024drivinggaussian}. Using LiDAR point as position prior, the distribution of 3D Gaussian can be represented as:
\begin{align}
    f(\mathbf{x}|\mu^{\scriptscriptstyle \text{L}}, \Sigma) = e^{-\frac{1}{2}(\mathbf{x}-\mu^{\scriptscriptstyle 
 \text{L}})^{\mathrm{T}}\Sigma^{-1}(\mathbf{x}-\mu^{\scriptscriptstyle  \text{L}})}
\end{align}
where $\mu^{\scriptscriptstyle  \text{L}}\in\mathbb{R}^{3}$ is the position of the LiDAR point, $\Sigma\in\mathbb{R}^{3\times3}$ is the covariance matrix of the 3D Gaussian.

To fully utilize the spatio-temporal consistency between different modalities during the Gaussian initialization, we employ RGB images to perform LiDAR point cloud coloring. This approach provides a prior for initializing the spherical harmonic coefficients of the Gaussians. To obtain accurate correspondences between LiDAR points $(x^{\scriptscriptstyle \text{L}},y^{\scriptscriptstyle \text{L}},z^{\scriptscriptstyle \text{L}})^{\mathrm{T}}\in\mathbb{R}^{3}$ and pixels $(u,v)^{\mathrm{T}}\in\mathbb{R}^{2}$, we segment the LiDAR points that fall within the frustum of each training view and subsequently project these points onto the pixel coordinate of the corresponding image to obtain RGB values:
\begin{align}
    C^{p_{i}^{\scriptscriptstyle \text{L}}}_{\text{rgb}}=\mathrm{Interpolate}\left(I,K_{\text{intr}}(Rp_{i}^{\scriptscriptstyle \text{L}}+t)\right),p_{i}^{\scriptscriptstyle \text{L}}\in F
\end{align}
where $C^{p_{i}^{\scriptscriptstyle \text{L}}}_{\text{rgb}}$ is the corresponding color of LiDAR point $p_{i}^{\scriptscriptstyle \text{L}}$, $I$ represents the image, $K_{\text{intr}}$ denotes the intrinsic parameters, while $R$ and $t$ represent the extrinsic parameters, associated with the camera corresponding to the image $I$, while $F$ denotes the set of LiDAR points within the frustum of the camera.

In addition, we filter the ground points from the LiDAR point cloud and employ 3D annotations for object bounding box erasing, in order to ensure high-quality reconstruction of the static background.


\textbf{Overfitting mitigation.} Unlike bounded scenarios that the vanilla 3D-GS can trivially render, autonomous driving scenes present challenges due to their boundlessness and sparse distribution of training views. This scarcity of supervision signals results in overfitting of training views, leading to floating artifacts and misalignment of geometric structures.

An important cause of overfitting is the confusion between near and distant scenes. Due to insufficient geometric information on distant landscapes, the Gaussians are prone to fit distant scenes as floating artifacts in near scenes during the training process, leading to background collapse. Referring to the strategy employed in~\cite{wu2024hgsmappingonlinedensemapping} for sky reconstruction, we mitigate this effect by adding spherical $\mathcal{P}_{\text{s}}$, composed of a set of points uniformly distributed along the periphery of the LiDAR point cloud. This operation aims to enhance the reconstruction quality of distant scenes beyond the LiDAR coverage. The spherical is also colored through multi-view RGB images to serve as the initial Gaussian prior.

\textbf{Mixed masking mechanism.} In autonomous driving scenes, there are environmental features that exhibit instability over time and contain less valuable information for place recognition. Therefore, we propose the mixed masking mechanism focusing on reconstructing only the stable parts during the Gaussian optimization process.

In light of the varying nature of unstable environmental features, we categorize the masked regions based on semantic information into static masks (e.g., sky and road surfaces) and dynamic masks (e.g., vehicles and pedestrians). We utilize a pre-trained Mask2Former~\cite{cheng2022masked} semantic segmentation network and 3D annotations to generate these two types of masks. The static masks are generated based on the semantic segmentation results. Areas of the training images covered by the static masks are overlaid with the background color of the 3D-GS renderer, serving to restrict the generation of Gaussians. The regions covered by the dynamic masks are generated through a two-step process. Firstly, 3D annotations are projected on images to obtain 2D bounding boxes. Then, pixels within the 2D bounding boxes that have the same semantic categories as the 3D annotations are selected, ensuring that static background areas are minimally masked. As directly culling the shadow areas of these dynamic objects may result in unnecessary information loss, we adopt a loss detach strategy, omitting the gradients for the masked regions during the Gaussian optimization process. This strategy mitigates the negative effects of dynamic objects and simultaneously maintains enough supervision for large-scale reconstruction.

\begin{figure}[t]
  \centering
  \includegraphics[width=1\linewidth]{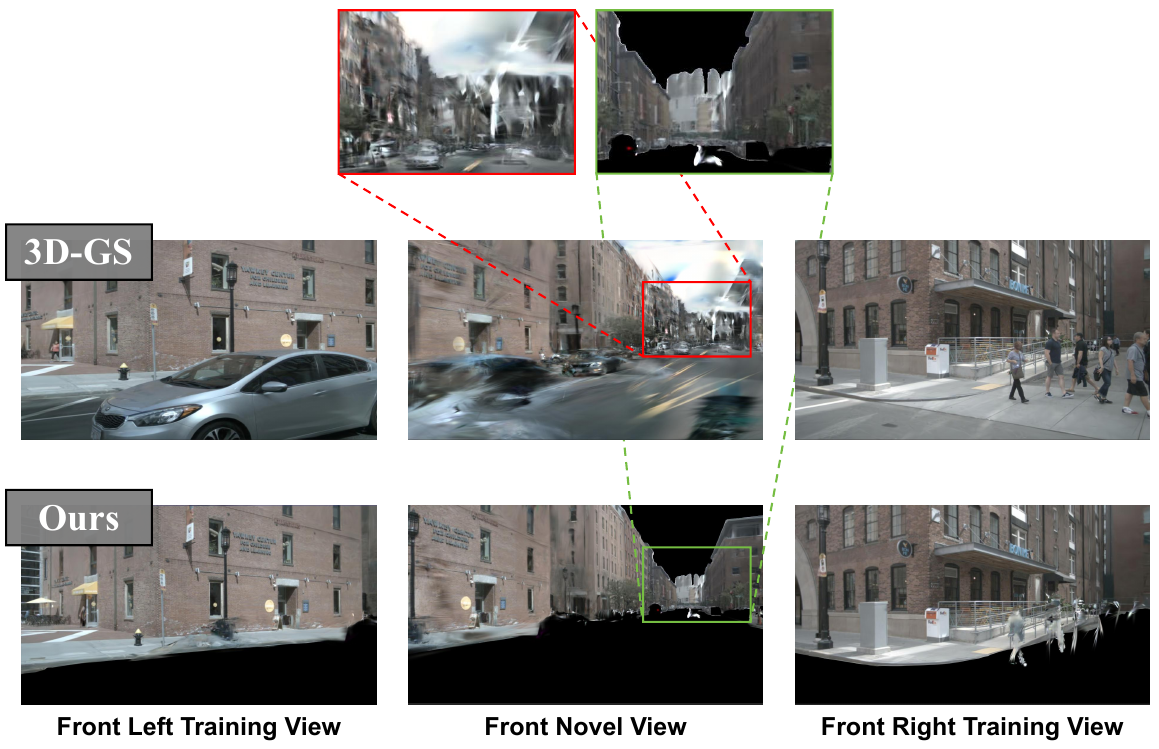}
  \caption{A comparison of the rendering results between our MGS and the vanilla 3D-GS. The environmental features of lesser significance for place recognition are masked, while the integration of LiDAR prior enhances the geometric accuracy of explicit scene reconstruction.}
  \label{fig:render}
  \vspace{-0.6cm}
\end{figure}

As demonstrated in~\figref{fig:render}, our proposed mixed masking mechanism effectively masks out unstable features. Additionally, the employment of LiDAR prior and the adaption of overfitting mitigation techniques contribute to maintaining a consistent scale and accurate geometric structure of the reconstructed scene. Consequently, our MGS effectively reconstructs Gaussian scenes suitable for place recognition.

\subsection{Global Descriptor Generator}
\label{sec:GDG}
Global Descriptor Generator is used to extract distinctive global descriptors from the proposed MGS representations. To extract the high-level spatio-temporal features, we first voxelize the MGS scene, and then extract local and global features through a backbone network composed of 3D graph convolutions~\cite{lin2020convolution} and transformer~\cite{vaswani2017attention} module. Finally, the spatio-temporal features are fed into NetVLAD-MLPs combos~\cite{uy2018pointnetvlad} and aggregated into discriminative descriptors.

\textbf{Voxel partition and encoding.} To tackle the disordered distribution of Gaussians, we first organize the MGS scene into a form that facilitates feature extraction through voxelization. Denote $\mathcal{G}=\{g_{m}=[x_{m}^{\scriptscriptstyle \text{G}},y_{m}^{\scriptscriptstyle \text{G}},z_{m}^{\scriptscriptstyle \text{G}},s_{m},q_{m},sh_{m},\alpha_{m}]^{\mathrm{T}}\in\mathbb{R}^{59}\}_{m=1...M}$ as a MGS scene, where $g_{m}$ represents the $m$-th Gaussian in the scene, $x_{m}^{\scriptscriptstyle \text{G}},y_{m}^{\scriptscriptstyle \text{G}},z_{m}^{\scriptscriptstyle \text{G}}$ denote the position of the Gaussian, $s_{m}$ means the scale matrix, $q_{m}$ is the quaternion, $sh_{m}$ means the SH coefficents, and $\alpha_{m}$ denotes the opacity. Inspired by~\cite{zhou2020cylinder3d}, we subdivide the space into voxels in cylindrical coordinates, to ensure the uniformity of the partitioning of the Gaussian scene. Subsequently, we allocate the Gaussians to the corresponding voxels through voxel partitioning, converting the Gaussian model with sizes of $M \times 59$ to $N \times H \times 59$, where $N$ is the number of voxels, and $H$ is the maximum number of Gaussians within each voxel.

Let $\mathcal{V}=\{g_{h}=[x_{h}^{\scriptscriptstyle \text{G}},y_{h}^{\scriptscriptstyle \text{G}},z_{h}^{\scriptscriptstyle \text{G}},s_{h},q_{h},sh_{h},\alpha_{h}]^{\mathrm{T}}\in\mathbb{R}^{59}\}_{h=1...H}$ as a non-empty voxel containing $H$ Gaussians. Inspired by~\cite{yan2018second}, we encode the voxel features by computing the mean of each attribute of the Gaussians within the voxel, to ensure the real-time performance and usability of the network. After the voxel encoding operation, the voxel set of shape $N \times H \times 59$ is encoded into an input form of $N \times 59$. We denote the encoded MGS scene representation as $\overline{\mathcal{G}}=\{\overline{g}_{n}=[\overline{x}_{n}^{\scriptscriptstyle \text{G}},\overline{y}_{n}^{\scriptscriptstyle \text{G}},\overline{z}_{n}^{\scriptscriptstyle \text{G}},\overline{s}_{n},\overline{q}_{n},\overline{sh}_{n},\overline{\alpha}_{n}]^{\mathrm{T}}\in\mathbb{R}^{59}\}_{n=1...N}$. Ultimately, voxel downsampling brings order to the Gaussian scene and reduces the computational burden.

\textbf{3D graph convolution.}  Inspired by the successful application of graph convolution in place recognition~\cite{liu2019lpd, hui2021pyramid}, we use a 3D-GCN-based~\cite{lin2020convolution} graph convolution backbone network to fully exploit the local features in the scene.

Based on the encoded MGS scene representation $\overline{\mathcal{G}}$, we construct a Gaussian graph according to the spatial relationships within it, using $p_{\text{gs}}=\{\overline{x}^{\scriptscriptstyle \text{G}},\overline{y}^{\scriptscriptstyle \text{G}},\overline{z}^{\scriptscriptstyle \text{G}}\}\in\mathbb{R}^{3}$ as the graph node's coordinate and $\boldsymbol{f}(p_{\text{gs}})=\{\overline{s},\overline{q},\overline{sh},\overline{\alpha}\}\in\mathbb{R}^{56}$ as the node's feature vector. To extract the local features of each node $p_{\text{gs}}^{n}$, we use kNN to construct the receptive field $R_{n}^{J}$ of $p_{\text{gs}}^{n}$ in 3D graph structure:
\begin{align}
    R_{n}^{J}=\left\{p_{\text{gs}}^{n},p_{\text{gs}}^{j}|\forall p_{\text{gs}}^{j}\in\mathcal{N}\left(p_{\text{gs}}^{n},J\right)\right\}
\end{align}
where $\mathcal{N}(.)$ denote the nearest neighbors operation using Euclidean distance, $J$ means the predefined number of neighbors, and $p_{\text{gs}}^{j}$ is the $j$-th neighbor of $p_{\text{gs}}^{n}$.

Additionally, we follow the definition in 3D-GCN, representing the 3D graph convolution kernel $K^{S}$ as a combination of unit support vectors with the origin as the starting point and their associated weights:
\begin{align}
    K^{S}=\left\{\boldsymbol{w}(k_{\text{C}}),\left(k_{s},\boldsymbol{w}(k_{s})\right)|s=1,2,...S\right\}
\end{align} 
where $k_{\text{C}}$ is the center of the kernel, $k_{s}$ are the support vectors, $\boldsymbol{w}(k)$ denote the associated weights, and $S$ is the number of the support vectors in the kernel. Thus, we can define the 3D-GCN graph convolution operation $Conv(R_{n}^{J},K^{S})$ as:
\begin{align}
\begin{split}
    Conv(R_{n}^{J},K^{S})&=\langle\boldsymbol{f}(p_{\text{gs}}^{n}),\boldsymbol{w}(k_{\text{C}})\rangle \\ &+\sum\limits_{s=1}^{S}\mathop{\max}_{j\in(1,J)}\left\{sim(p_{\text{gs}}^{j},k_{s})\right\}
\end{split}
\end{align}
\begin{align}
    sim(p_{\text{gs}}^{j},k_{s})=\langle\boldsymbol{f}(p_{\text{gs}}^{j}),\boldsymbol{w}(k_{s})\rangle\frac{\langle d_{j,n},k_{s}\rangle}{||d_{j,n}||\cdot||k_{s}||}
\end{align}

We perform zero-mean normalization on the coordinates of the Gaussian graph and subsequently feed the Gaussian graph into stacked 3D graph convolution layers, 3D graph max-pooling layers~\cite{lin2020convolution}, and ReLU nonlinear activation layers. The graph convolution backbone network generates output feature graph $F^{\text{out}}\in\mathbb{R}^{B \times N_{\text{out}} \times CH}$ based on the input features of Gaussian graph $F^{\text{in}}\in\mathbb{R}^{B \times N \times 56}$, which are then used for subsequent processing, where $B$ means the batch size, and $CH$ denotes the output channel dimension. The use of graph convolution enhances the network's ability to aggregate local spatio-temporal features within the Gaussian graph, contributing to the discriminativity of place recognition representations.

\begin{figure}[t]
  \centering
  \includegraphics[width=1\linewidth]{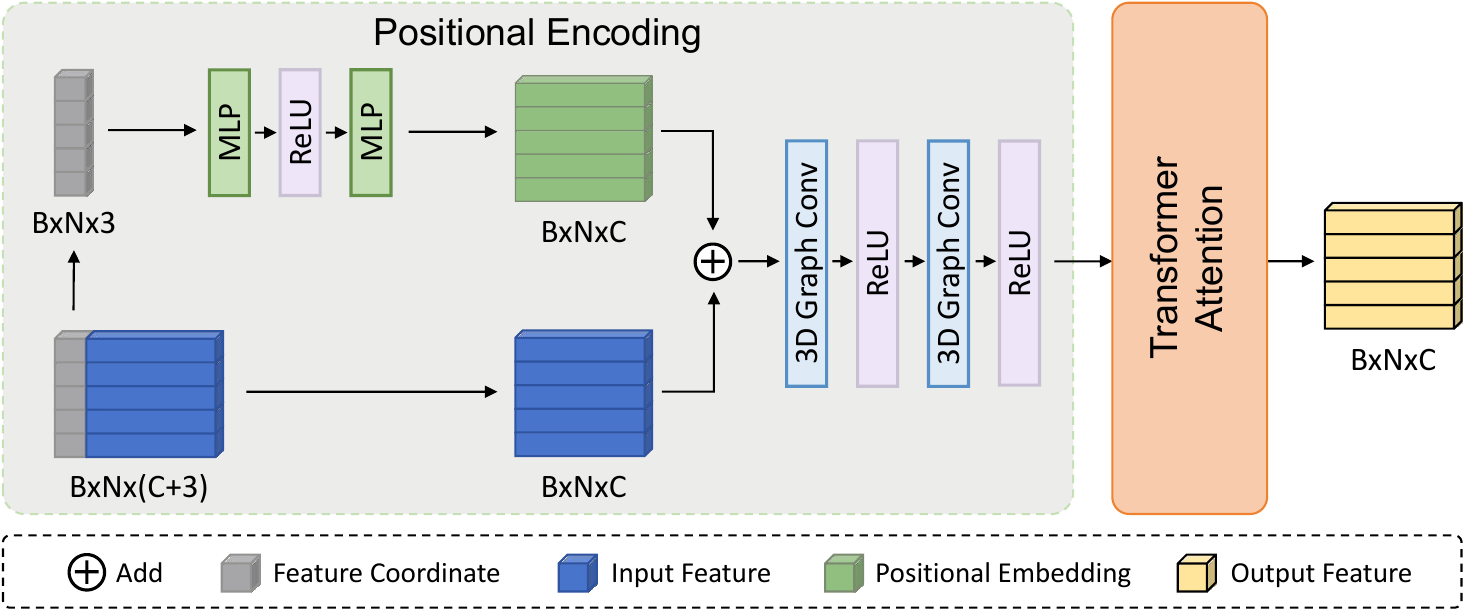}
  \caption{The detailed architecture of transformer module. Feature coordinates are explicitly encoded as positional embeddings and fused with features through graph convolutions. A transformer attention is used to extract global context from the features.}
  \label{fig:pos}
  \vspace{-0.6cm}
\end{figure}

\textbf{Transformer module.} Inspired by previous works~\cite{ma2022overlaptransformer, xu2024explicit}, we use transformers to extract the global context within the feature graph, to boost place recognition performance. 
The architecture of our devised transformer module is depicted in~\figref{fig:pos}. To enable the transformer to capture the spatial correlations embedded in the feature graph, we use a feed-forward network to encode the coordinates of the feature graph $p_{\mathrm{feat}}^{i}$ into learnable positional embeddings. We add the positional embeddings to the features and use stacked 3D graph convolution layers for feature fusion. Then we feed the position-encoded features into multi-head attention to fully extract the global spatio-temporal information in the scene. The self-attention mechanism can be formulated as:
\begin{align}
    \mathcal{A}=\mathrm{Attention}(Q,K,V)=\mathrm{softmax}\left(\frac{QK^{\mathrm{T}}}{\sqrt{d_{\text{k}}}}\right)V
\end{align}
where $\mathcal{A}$ denotes the feature with global context, $Q,K,V$ represent the queries, keys and values respectively, and $d_{\text{k}}$ is the dimension of keys.

\begin{table*}[ht]
  \centering
  \begin{center}
  	\setlength{\tabcolsep}{7pt}
  	\renewcommand\arraystretch{1.0}
    \caption{Comparision of place recognition performance on the BS, SON, and SQ splits}
    \vspace{-0.2cm}
    \footnotesize{
        \begin{tabular}{cccccccccccc}
          \toprule
          \multirow{2}{*}{Methods} & \multirow{2}{*}{Sequence$^1$} & \multirow{2}{*}{Modality$^2$} &\multicolumn{3}{c}{BS split}&\multicolumn{3}{c}{SON split}&\multicolumn{3}{c}{SQ split} \\ 
          \cline{4-12}
          \addlinespace[3pt]
         ~&~&~ & AR@1 &AR@5 & AR@10 & AR@1 & AR@5 & AR@10 & AR@1 & AR@5 &  AR@10 \\ \midrule
          AnyLoc~\cite{keetha2023anyloc} & $\times$ & V & 80.79 & 89.76 & 94.11 & 97.47 & 98.74 & \textbf{100.00} & 90.55 & 92.07 & 93.29 \\
          OT~\cite{ma2022overlaptransformer} & $\times$ & L & 67.60 & 82.75 & 86.96 & 92.68 & 97.22 & 98.23 & 96.95 & 99.39 & 99.39 \\
          MinkLoc++~\cite{komorowski2021minkloc++} & $\times$ & V+L & 74.19 & 90.04 & 92.99 & 86.62 & 96.46 & 97.98 & 88.11 & 94.21 & 95.12 \\
          LCPR~\cite{zhou2023lcpr} & $\times$ & V+L & 89.48 & 96.21 & 97.34 & 96.46 & 99.24 & 99.49 & 90.85 & 97.87 & 98.48 \\
          SeqNet~\cite{garg2021seqnet} & $\checkmark$ & V & 74.86 & 83.29 & 87.78 & 87.09 & 92.66 & 95.19 & 78.59 & 86.85 & 88.99 \\
          SeqOT~\cite{ma2022seqot} & $\checkmark$ & L & 78.12 & 88.78 & 92.01 & 97.47 & 98.48 & 98.99 & 98.78 & 99.39 & 99.39 \\
          Autoplace~\cite{cait2022autoplace} & $\checkmark$ & R & 83.85 & 93.12 & 95.93 & 95.70 & 98.73 & 99.24 & 95.72 & 98.78 & 98.78 \\
          GSPR-L (ours) & $\checkmark$ & V+L & 96.05 & 99.16 & 99.30 & 93.94 & 97.98 & 99.24 & 94.21 & 98.17 & 99.70 \\
          GSPR (ours) & $\checkmark$ & V+L & \textbf{98.74} & \textbf{99.44} & \textbf{99.44} & \textbf{98.99} & \textbf{99.75} & \textbf{100.00} & \textbf{99.09} & \textbf{99.70} & \textbf{99.70} \\
          \bottomrule           
          \addlinespace[1pt]
        \multicolumn{12}{p{0.9\linewidth}}{$^1$ Use sequential data. $^2$ V: Visual, L: LiDAR, R: Radar, V+L: Visual+LiDAR.}\\
        \end{tabular}
        }
    \label{tab:bs_son_sq}
    \end{center}
    \vspace{-0.7cm}
\end{table*}

\begin{table*}[ht]
  \centering
  \begin{center}
  	\vspace{0.3cm}
  	\setlength{\tabcolsep}{7pt}
  	\renewcommand\arraystretch{1.0}
    \caption{Comparision of place recognition performance on the KITTI and KITTI-360 datasets}
    \vspace{-0.2cm}
    \footnotesize{
        \begin{tabular}{cccccccccccc}
          \toprule
          \multirow{2}{*}{Methods} & \multirow{2}{*}{Sequence$^1$} & \multirow{2}{*}{Modality$^2$} &\multicolumn{3}{c}{KITTI}&\multicolumn{3}{c}{KITTI-360}&\multicolumn{3}{c}{Generalization} \\ \cline{4-12} 
        \addlinespace[3pt]
         ~&~&~ & AR@1 &AR@5 & AR@10 & AR@1 & AR@5 & AR@10 & AR@1 & AR@5 &  AR@10 \\ \midrule
          MinkLoc++~\cite{komorowski2021minkloc++} & $\times$ & V+L & 95.76 & 99.15 & 99.72 & 95.89 & 99.03 & 99.27 & 96.37 & 99.27 & 99.52 \\
          SeqOT~\cite{ma2022seqot} & $\checkmark$ & L & 97.46 & 99.15 & 99.72 & 98.07 & 99.27 & 99.40 & 98.31 & 99.52 & \textbf{99.88} \\
          GSPR-L (ours) & $\checkmark$ & V+L & 98.31 & 99.72 & 99.72 & 98.55 & 99.40 & 99.40 & 97.34 & 99.15 & 99.27 \\
          GSPR (ours) & $\checkmark$ & V+L & \textbf{99.44} & \textbf{99.72} & 
          \textbf{100.00} & \textbf{99.15} & \textbf{99.64} & \textbf{99.64} & \textbf{98.91}& \textbf{99.64} & 99.64 \\
          \bottomrule
        \addlinespace[1pt]
        \multicolumn{12}{p{0.9\linewidth}}{$^1$ Use sequential data. $^2$ V: Visual, L: LiDAR, R: Radar, V+L: Visual+LiDAR.}\\
        \end{tabular}
        }
    \label{tab:kitti}
    \end{center}
    \vspace{-0.9cm}
\end{table*}

\subsection{Two-step Training Strategy}
\label{sec:network_training}
We adopt a two-stage process to train the GSPR. Firstly, we train explicit representations of autonomous driving scenes based on Multimodal Gaussian Splatting. Subsequently, the Global Descriptor Generator is trained for place recognition using the generated MGS scene representations.

Following 3D-GS~\cite{kerbl20233d}, we supervise the Gaussian optimization process of Multimodal Gaussian Splatting using the combination of the Mean Absolute Error loss $\mathcal{L}_{1}$ and the  Structural Similarity Index Measure loss $\mathcal{L}_{\text{D-SSIM}}$:
\begin{align}\label{eq10}
\begin{split}
    \mathcal{L}_{\text{MGS}}\left(I_{\text{render}},I_{\text{gt}}\right)&=(1-\lambda)\mathcal{L}_{1}\left(I_{\text{render}},I_{\text{gt}}\right) \\&+\lambda\mathcal{L}_{\text{D-SSIM}}\left(I_{\text{render}},I_{\text{gt}}\right)
\end{split}
\end{align}
where $I_{\text{render}}$ is the image rendered by the 3D-GS renderer from the MGS scene representation, and $I_{\text{gt}}$ represents the ground-truth image. We use $\mathcal{L}_{\text{MGS}}$ to supervise the iterative refinement of MGS scene representations to accurately reconstruct the autonomous driving scenarios.

To supervise the Global Descriptor Generator, we employ the contrastive learning scheme. For each query descriptor $\mathcal{D}$ representing an MGS scene representation, we choose $k_{\text{pos}}$ positive descriptors $\left\{\mathcal{D}_{\text{pos}}\right\}$ and $k_{\text{neg}}$ negative descriptors $\left\{\mathcal{D}_{\text{neg}}\right\}$ to construct a triplet $\mathcal{T}=\left(\mathcal{D},\left\{\mathcal{D}_{\text{pos}}\right\},\left\{\mathcal{D}_{\text{neg}}\right\}\right)$. Following previous works~\cite{cait2022autoplace, zhou2023lcpr}, we define samples within 9\,meters of the query sample as positive, otherwise negative. We input the triplets into lazy triplet loss~\cite{uy2018pointnetvlad} to compute the loss, accelerating network convergence and boosting place recognition performance through mini-batch hard mining. The loss function is given by:
\begin{align}
    \hspace{-0.15cm} \mathcal{L}_{\text{GDG}}(\mathcal{T})\!=\!\left[\beta\!+\!\mathop{\min}_{o}(d(\mathcal{D}\!,\mathcal{D}_{\text{pos}}^{o}))\!-\!\mathop{\max}_{a}(d(\mathcal{D}\!,\mathcal{D}_{\text{neg}}^{a}))\right]_+
\end{align}
where $\left [ \cdots \right ]_+$ denotes the hinge loss, $d(\cdot)$ is the Euclidean distance between a pair of descriptors, and $\beta$ is the margin.

\section{Experiments}
\label{sec:experiments}

\subsection{Experimental Setup}
\label{sec:expe_setup}
We use three datasets, nuScenes~\cite{caesar2020nuscenes}, KITTI~\cite{geiger2013vision}, and KITTI-360~\cite{liao2022kitti}, to evaluate the place recognition accuracy and generalization performance of our proposed GSPR.

\textbf{nuScenes.} It includes autonomous driving scenes collected from four different locations: Boston Seaport (BS), SG-OneNorth (SON), SG-Queenstown (SQ), and SG-HollandVillage (SHV). It provides multimodal data from 32-beam LiDAR and multi-view cameras. To obtain statistically significant results, we conduct experiments on the BS, SON, and SQ splits, which have sufficient loop closures and diverse situations. Our data preparation pipeline mainly follows~\cite{cait2022autoplace, zhou2023lcpr}. Inspired by~\cite{xu2023ring++}, we construct a sparse scan map by downsampling the \textit{database set} at 3\,meter intervals and the \textit{test set} at 9\,meter intervals, to evaluate recognition accuracy under large viewpoint differences. To construct a sequence, we use the current observation, along with the previous and next observations that are temporally adjacent to it.

\textbf{KITTI.} It is a standard autonomous driving dataset that includes various urban scenarios and traffic conditions. It provides multimodal data, including front-view stereo cameras and the Velodyne HDL-64E LiDAR, with GNSS-based ground-truth poses. We select Sequence 02 for training and Sequence 00 for testing, and perform data partitioning in the same manner as for the nuScenes dataset.

\textbf{KITTI-360.} It is a larger multimodal dataset compared to KITTI, with a similar sensor configuration. We select 2013-05-28-drive-0000 for training and 2013-05-28-drive-0002 for testing. Additionally, we transfer the weights trained on the KITTI dataset to KITTI-360 to evaluate the cross-dataset generalization ability of our proposed method. Notably, the field of view of the cameras in KITTI and KITTI-360 is considerably smaller than that of nuScenes, making the panoramic reconstruction results unavailable. We demonstrate the robustness of GSPR to sensor configuration through experiments on KITTI and KITTI-360.

\subsection{GSPR Implementation Details}
\label{sec:detail}
For the Gaussian Optimization module, we set the training iteration to 400, which significantly accelerates the training and inference of the network. This setting sacrifices some rendering quality, but is sufficient to reconstruct the dense and uniform Gaussian scene representations. For the 3D graph convolution backbone, we set the number of neighbors $J=25$, the kernel support number $S=1$, and the sampling rate of the 3D graph max-pooling $r_{\text{pool}}=0.25$. For the transformer module, we set the positional embedding and the feature embedding dimension $d_{\text{pe}}=d_{\text{model}}=512$, the feed-forward dimension $d_{\text{ffn}}=1024$, and the number of heads $n_{\text{head}}=8$. For the NetVLAD module, we set the number of clusters $d_{\text{cluster}}=64$, and the descriptor dimension $d_{\text{out}}=256$. An ADAM optimizer is used to train the network, while the initial learning rate is set to $1 \times 10^{-5}$ and decays by a factor of 0.5 every 5 epochs. For the lazy triplet loss, we set the number of positive samples $k_{\text{pos}}=2$, the number of negative samples $k_{\text{neg}}=6$, and the margin $\beta=0.5$. 

In addition, we set the number of input voxels in GSPR to $N=4096/8192$ during training/inference respectively. We also design a lightweight version, GSPR-L, with only half the number of input voxels in GSPR for inference.
All experiments are conducted on a system with an Intel i7-14700KF CPU and an Nvidia RTX 4060Ti GPU.

\subsection{Evaluation for Place Recognition}
\label{sec:eval_res}
To validate the place recognition performance of GSPR in large-scale outdoor environments, we compare it with state-of-the-art baseline methods, including the visual-based methods AnyLoc~\cite{keetha2023anyloc} and SeqNet~\cite{garg2021seqnet}, the LiDAR-based methods OverlapTransformer~\cite{ma2022overlaptransformer}, and SeqOT~\cite{ma2022seqot}, the radar-based method Autoplace~\cite{cait2022autoplace}, and the multimodal methods MinkLoc++~\cite{komorowski2021minkloc++} and LCPR~\cite{zhou2023lcpr}. Among these, SeqNet~\cite{garg2021seqnet}, SeqOT~\cite{ma2022seqot}, and Autoplace~\cite{cait2022autoplace} use sequential observations as inputs, compared to the other baselines only using one single frame for each retrieval.
We try to reproduce the baselines using their open source code. During the mixed masking process of GSPR, we use ground-truth 3D annotations for training on nuScenes. For inference on nuScenes and the entire deployment on KITTI and KITTI-360, we use annotations generated by PointPillars~\cite{lang2019pointpillars}. We set the sequence length for all sequence-enhanced place recognition methods to 3 for fairness.

Following previous works~\cite{cait2022autoplace, ma2023cvtnet}, we use average top 1 recall (AR@1), top 5 recall (AR@5),
and top 10 recall (AR@10) as metrics to evaluate the place recognition performance. The results on the nuScenes dataset are shown in~\tabref{tab:bs_son_sq}. In addressing challenging scenarios that include rain and nighttime conditions, our proposed GSPR holds the best recognition accuracy on all metrics, while GSPR-L strikes a balance between inference speed (approximately one-third of the GSPR runtime for global descriptor generation) and recognition accuracy. This demonstrates that our method effectively handles scenarios where unimodal approaches fail, and achieves good recognition accuracy under large viewpoint differences.

The experimental results on the KITTI and KITTI-360 datasets are shown in~\tabref{tab:kitti}. Our method presents good place recognition accuracy on both KITTI and KITTI-360 datasets, while showing strong generalization performance in cross-dataset scenarios. Furthermore, the experimental results on KITTI and KITTI-360 also demonstrate the solid robustness of GSPR in the case where panoramic reconstruction results are not available. When the 3D-GS reconstruction is supervised using only a front-view stereo camera, GSPR still maintains the best recognition accuracy.

\begin{table}[t]
    \centering
    \tabcolsep=1.2mm
    \renewcommand\arraystretch{0.9}
    \caption{Ablation study of improvement strategies}
    \vspace{-0.2cm}
    \footnotesize{
        \begin{tabular}{cccc|ccc}
            \toprule
             \makecell{LiDAR \\ Initialization} & \makecell{Spherical \\ Dome} & \makecell{Static \\ Mask} & \makecell{Dynamic \\ Mask} & AR@1 & AR@5 & AR@10 \\ \midrule
               &  &  &  & 12.81 & 28.77 & 39.02 \\
             \checkmark &  &  &  & 91.85 & 98.04 & 98.88 \\
             \checkmark & \checkmark &  &  & 92.13 & 98.46 & 99.16 \\
             \checkmark & \checkmark & \checkmark &  & 93.67 & 98.88 & 99.02  \\
             \checkmark & \checkmark & \checkmark & \checkmark & \textbf{96.05} & \textbf{99.16} & \textbf{99.30} \\
             \bottomrule
        \end{tabular} 
    }
    \label{tab:trick_ablation}
    \vspace{-0.6cm}
\end{table}

\subsection{Ablation Studies}
\label{sec:eval_improvement}
\textbf{Improvement strategies.} We ablate the improvement strategies of our MGS module in generating Gaussian scenes tailored for place recognition. The experimental results of GSPR-L in~\tabref{tab:trick_ablation} show that each improvement strategy of MGS has a positive effect on place recognition performance. In particular, using the Gaussian scenes generated by the vanilla 3D-GS as input results in a relatively low recognition accuracy (the first row of~\tabref{tab:trick_ablation}). This is probably due to the difficulty of SfM in producing reliable sparse reconstruction results on the nuScenes dataset, resulting in poor Gaussian initialization and suboptimal scene reconstruction.

\textbf{Input features.} A Gaussian $g=[\mu^{\scriptscriptstyle \text{G}},s,q,sh,\alpha]\in\mathbb{R}^{59}$ is composed of different parts of features, including position $\mu^{\scriptscriptstyle \text{G}}\in\mathbb{R}^{3}$, scale $s\in\mathbb{R}^{3}$, rotation $q\in\mathbb{R}^{4}$, SH coefficients $sh\in\mathbb{R}^{48}$, and opacity $\alpha\in\mathbb{R}^{1}$. We ablate these input features using the BS split to assess their impact on recognition performance. The results of GSPR-L shown in~\tabref{tab:input_ablation} indicate that, in addition to the SH coefficients, position, scale, and opacity are crucial for place recognition, while the rotation feature contributes less. This suggests that ``where it is'' is more expressive than ``which direction it is heading'' for Gaussian-based place description, as the former corresponds more directly to the explicit spatial structure of the places.

\begin{table}
    \centering
    \tabcolsep=1.2mm
    \renewcommand\arraystretch{0.9}
    \caption{Ablation study of input features on the BS split}
    \vspace{-0.2cm}
    \begin{tabular}{ccccc|ccc}
        \toprule
         SH & Opacity & Rotation & Scale & Position & AR@1 & AR@5 & AR@10 \\ \midrule
         \checkmark &  &  &  &  & 73.35 & 88.36 & 91.44 \\
         \checkmark & \checkmark &  &  &  & 82.47 & 94.53 & 97.19 \\
         \checkmark & \checkmark & \checkmark &  &  & 83.73 & 93.83 & 96.21  \\
         \checkmark & \checkmark & \checkmark & \checkmark &  & 91.30 & 98.18 & 98.88 \\
         \checkmark & \checkmark & \checkmark & \checkmark & \checkmark & \textbf{96.05} & \textbf{99.16} & \textbf{99.30}  \\    
         \bottomrule
    \end{tabular} 
    \label{tab:input_ablation}
    \vspace{-0.6cm}
\end{table}

\begin{figure}
\vspace{0.3cm}
  \centering
  \includegraphics[width=1\linewidth]{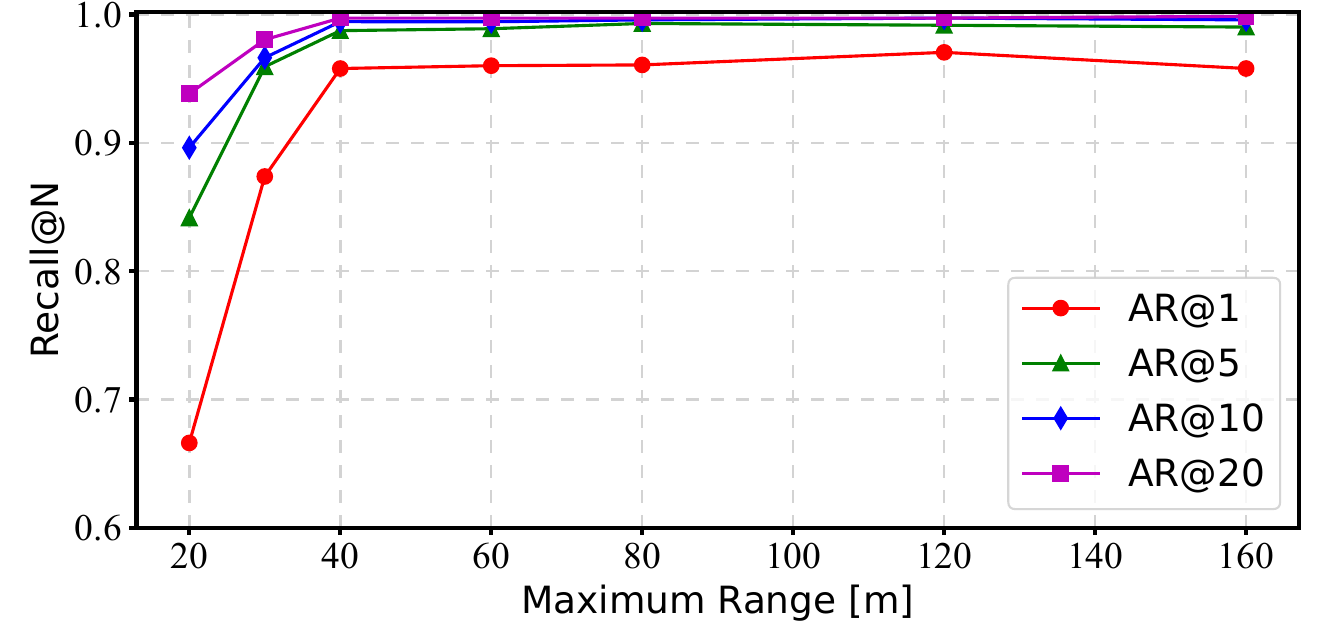}
  \vspace{-0.5cm}
  \caption{The impact of maximum sampling distance on place recognition performance.}
  \label{fig:range}
  \vspace{-0.3cm}
\end{figure}

\textbf{Maximum sampling ranges.} We further explore the impact of varying the maximum sampling range during voxel partitioning on GSPR's recognition performance, focusing on the contribution of Gaussians distributed at different distances within the scene. As shown in~\figref{fig:range}, the best place recognition performance occurs when the maximum sampling distance is at least 40\,meters. Notably, the AR@1 does not significantly increase with sampling range increase after 40\,meters. A possible reason is that each Gaussian scene is initialized from LiDAR data where points at greater distances are more sparse, leading to less distinct spatio-temporal features to boost the recognition performance.

\section{Conclusion}
\label{sec:conclusion}
In this paper, we present GSPR, a novel multimodal place recognition network based on 3D-GS. Our method proposes Multimodal Gaussian Splatting to harmonize multi-view RGB images and LiDAR point clouds into a unified spatio-temporal MGS scene representation tailored for place recognition. To manage the unordered Gaussians, we apply voxel downsampling for efficient data organization. We further propose using 3D graph convolution networks and transformer module to exploit local and global spatio-temporal features from Gaussian graphs, generating discriminative global  descriptors. Experimental results indicate that our method outperforms state-of-the-art baselines, demonstrating the advantages of the 3D-GS-based multimodal fusion approach for challenging place recognition tasks.


\bibliographystyle{unsrt}

\footnotesize{
\bibliography{glorified, new}}

\end{document}

%% file: stachnisslab-latex.tex
\usepackage{graphics}           
\usepackage{times}              
\usepackage{amsmath}            
\usepackage{amssymb}            
\usepackage{graphicx}
\usepackage{algorithm}
\usepackage[noend]{algpseudocode}
\usepackage{booktabs}
\usepackage{color}
\definecolor{instructioncolor}{rgb}{.5,.5,.5}

\usepackage[font=small]{caption}

\def\figref#1{Fig.~\ref{#1}}
\def\tabref#1{Tab.~\ref{#1}}
\def\eqref#1{Eq.~(\ref{#1})}

\makeatletter
\usepackage{xspace}
\DeclareRobustCommand\onedot{\futurelet\@let@token\@onedot}
\def\@onedot{\ifx\@let@token.\else.\null\fi\xspace}

\makeatother

\usepackage{array}
\newcolumntype{L}[1]{>{\raggedright\let\newline\\\arraybackslash\hspace{0pt}}m{#1}}
\newcolumntype{C}[1]{>{\centering\let\newline\\\arraybackslash\hspace{0pt}}m{#1}}
\newcolumntype{R}[1]{>{\raggedleft\let\newline\\\arraybackslash\hspace{0pt}}m{#1}}










%% file: GSPR.bbl
\begin{thebibliography}{10}

\bibitem{arandjelovic2013all}
Relja Arandjelovic and Andrew Zisserman.
\newblock All about {VLAD}.
\newblock In {\em Proceedings of the IEEE conference on Computer Vision and Pattern Recognition}, pages 1578--1585, 2013.

\bibitem{arandjelovic2016netvlad}
Relja Arandjelovic, Petr Gronat, Akihiko Torii, Tomas Pajdla, and Josef Sivic.
\newblock {NetVLAD}: Cnn architecture for weakly supervised place recognition.
\newblock In {\em Proceedings of the IEEE conference on computer vision and pattern recognition}, pages 5297--5307, 2016.

\bibitem{keetha2023anyloc}
Nikhil Keetha, Avneesh Mishra, Jay Karhade, Krishna~Murthy Jatavallabhula, Sebastian Scherer, Madhava Krishna, and Sourav Garg.
\newblock {AnyLoc}: Towards universal visual place recognition.
\newblock {\em IEEE Robotics and Automation Letters}, 2023.

\bibitem{uy2018pointnetvlad}
Mikaela~Angelina Uy and Gim~Hee Lee.
\newblock {PointNetVLAD}: Deep point cloud based retrieval for large-scale place recognition.
\newblock In {\em Proceedings of the IEEE conference on computer vision and pattern recognition}, pages 4470--4479, 2018.

\bibitem{liu2019lpd}
Zhe Liu, Shunbo Zhou, Chuanzhe Suo, Peng Yin, Wen Chen, Hesheng Wang, Haoang Li, and Yun-Hui Liu.
\newblock {LPD-Net}: 3{D} point cloud learning for large-scale place recognition and environment analysis.
\newblock In {\em Proceedings of the IEEE/CVF International Conference on Computer Vision}, pages 2831--2840, 2019.

\bibitem{ma2022overlaptransformer}
Junyi Ma, Jun Zhang, Jintao Xu, Rui Ai, Weihao Gu, and Xieyuanli Chen.
\newblock {OverlapTransformer}: An efficient and yaw-angle-invariant transformer network for lidar-based place recognition.
\newblock {\em IEEE Robotics and Automation Letters}, 7(3):6958--6965, 2022.

\bibitem{komorowski2021minkloc++}
Jacek Komorowski, Monika Wysocza{\'n}ska, and Tomasz Trzcinski.
\newblock {MinkLoc++}: lidar and monocular image fusion for place recognition.
\newblock In {\em 2021 International Joint Conference on Neural Networks (IJCNN)}, pages 1--8. IEEE, 2021.

\bibitem{zhou2023lcpr}
Zijie Zhou, Jingyi Xu, Guangming Xiong, and Junyi Ma.
\newblock {LCPR}: A multi-scale attention-based lidar-camera fusion network for place recognition.
\newblock {\em IEEE Robotics and Automation Letters}, 2023.

\bibitem{kerbl20233d}
Bernhard Kerbl, Georgios Kopanas, Thomas Leimk{\"u}hler, and George Drettakis.
\newblock {3D} gaussian splatting for real-time radiance field rendering.
\newblock {\em ACM Trans. Graph.}, 42(4):139--1, 2023.

\bibitem{milford2012seqslam}
Michael~J Milford and Gordon~F Wyeth.
\newblock {SeqSLAM}: Visual route-based navigation for sunny summer days and stormy winter nights.
\newblock In {\em 2012 IEEE international conference on robotics and automation}, pages 1643--1649. IEEE, 2012.

\bibitem{9944848}
Yunge Cui, Xieyuanli Chen, Yinlong Zhang, Jiahua Dong, Qingxiao Wu, and Feng Zhu.
\newblock {BoW3D}: Bag of words for real-time loop closing in 3d lidar slam.
\newblock {\em IEEE Robotics and Automation Letters}, 8(5):2828--2835, 2023.

\bibitem{garg2020delta}
Sourav Garg, Ben Harwood, Gaurangi Anand, and Michael Milford.
\newblock {Delta Descriptors}: Change-based place representation for robust visual localization.
\newblock {\em IEEE Robotics and Automation Letters}, 5(4):5120--5127, 2020.

\bibitem{mereu2022learning}
Riccardo Mereu, Gabriele Trivigno, Gabriele Berton, Carlo Masone, and Barbara Caputo.
\newblock Learning sequential descriptors for sequence-based visual place recognition.
\newblock {\em IEEE Robotics and Automation Letters}, 7(4):10383--10390, 2022.

\bibitem{izquierdo2024optimal}
Sergio Izquierdo and Javier Civera.
\newblock Optimal transport aggregation for visual place recognition.
\newblock In {\em Proceedings of the IEEE/CVF Conference on Computer Vision and Pattern Recognition}, pages 17658--17668, 2024.

\bibitem{berton2023jist}
Gabriele Berton, Gabriele Trivigno, Barbara Caputo, and Carlo Masone.
\newblock {JIST}: Joint image and sequence training for sequential visual place recognition.
\newblock {\em IEEE Robotics and Automation Letters}, 2023.

\bibitem{hui2021pyramid}
Le~Hui, Hang Yang, Mingmei Cheng, Jin Xie, and Jian Yang.
\newblock Pyramid point cloud transformer for large-scale place recognition.
\newblock In {\em Proceedings of the IEEE/CVF International Conference on Computer Vision}, pages 6098--6107, 2021.

\bibitem{cait2022autoplace}
Kaiwen Cait, Bing Wang, and Chris~Xiaoxuan Lu.
\newblock {AutoPlace}: Robust place recognition with single-chip automotive radar.
\newblock In {\em 2022 International Conference on Robotics and Automation (ICRA)}, pages 2222--2228. IEEE, 2022.

\bibitem{ma2022seqot}
Junyi Ma, Xieyuanli Chen, Jingyi Xu, and Guangming Xiong.
\newblock {SeqOT}: A spatial-temporal transformer network for place recognition using sequential lidar data.
\newblock {\em IEEE Transactions on Industrial Electronics}, 2022.

\bibitem{ma2023cvtnet}
Junyi Ma, Guangming Xiong, Jingyi Xu, and Xieyuanli Chen.
\newblock {CVTNet}: A cross-view transformer network for lidar-based place recognition in autonomous driving environments.
\newblock {\em IEEE Transactions on Industrial Informatics}, 2023.

\bibitem{chen2022overlapnet}
Xieyuanli Chen, Thomas L{\"a}be, Andres Milioto, Timo R{\"o}hling, Jens Behley, and Cyrill Stachniss.
\newblock {OverlapNet}: A siamese network for computing lidar scan similarity with applications to loop closing and localization.
\newblock {\em Autonomous Robots}, pages 1--21, 2022.

\bibitem{luo2021bvmatch}
Lun Luo, Si-Yuan Cao, Bin Han, Hui-Liang Shen, and Junwei Li.
\newblock {BVMatch}: Lidar-based place recognition using bird's-eye view images.
\newblock {\em IEEE Robotics and Automation Letters}, 6(3):6076--6083, 2021.

\bibitem{luo2023bevplace}
Lun Luo, Shuhang Zheng, Yixuan Li, Yongzhi Fan, Beinan Yu, Si-Yuan Cao, Junwei Li, and Hui-Liang Shen.
\newblock {BEVPlace}: Learning lidar-based place recognition using bird's eye view images.
\newblock In {\em Proceedings of the IEEE/CVF International Conference on Computer Vision}, pages 8700--8709, 2023.

\bibitem{lai2022adafusion}
Haowen Lai, Peng Yin, and Sebastian Scherer.
\newblock {AdaFusion}: Visual-lidar fusion with adaptive weights for place recognition.
\newblock {\em IEEE Robotics and Automation Letters}, 7(4):12038--12045, 2022.

\bibitem{xu2024explicit}
Jingyi Xu, Junyi Ma, Qi~Wu, Zijie Zhou, Yue Wang, Xieyuanli Chen, and Ling Pei.
\newblock Explicit interaction for fusion-based place recognition.
\newblock {\em arXiv preprint arXiv:2402.17264}, 2024.

\bibitem{yan2024street}
Yunzhi Yan, Haotong Lin, Chenxu Zhou, Weijie Wang, Haiyang Sun, Kun Zhan, Xianpeng Lang, Xiaowei Zhou, and Sida Peng.
\newblock Street gaussians for modeling dynamic urban scenes.
\newblock {\em arXiv preprint arXiv:2401.01339}, 2024.

\bibitem{zhou2024drivinggaussian}
Xiaoyu Zhou, Zhiwei Lin, Xiaojun Shan, Yongtao Wang, Deqing Sun, and Ming-Hsuan Yang.
\newblock {DrivingGaussian}: Composite gaussian splatting for surrounding dynamic autonomous driving scenes.
\newblock In {\em Proceedings of the IEEE/CVF Conference on Computer Vision and Pattern Recognition}, pages 21634--21643, 2024.

\bibitem{huang2024textit}
Nan Huang, Xiaobao Wei, Wenzhao Zheng, Pengju An, Ming Lu, Wei Zhan, Masayoshi Tomizuka, Kurt Keutzer, and Shanghang Zhang.
\newblock {S3Gaussian}: Self-supervised street gaussians for autonomous driving.
\newblock {\em arXiv preprint arXiv:2405.20323}, 2024.

\bibitem{shi2024dhgs}
Xi~Shi, Lingli Chen, Peng Wei, Xi~Wu, Tian Jiang, Yonggang Luo, and Lecheng Xie.
\newblock {DHGS}: Decoupled hybrid gaussian splatting for driving scene.
\newblock {\em arXiv preprint arXiv:2407.16600}, 2024.

\bibitem{wu2024hgsmappingonlinedensemapping}
Ke~Wu, Kaizhao Zhang, Zhiwei Zhang, Shanshuai Yuan, Muer Tie, Julong Wei, Zijun Xu, Jieru Zhao, Zhongxue Gan, and Wenchao Ding.
\newblock {HGS-Mapping}: Online dense mapping using hybrid gaussian representation in urban scenes, 2024.

\bibitem{cheng2022masked}
Bowen Cheng, Ishan Misra, Alexander~G Schwing, Alexander Kirillov, and Rohit Girdhar.
\newblock Masked-attention mask transformer for universal image segmentation.
\newblock In {\em Proceedings of the IEEE/CVF conference on computer vision and pattern recognition}, pages 1290--1299, 2022.

\bibitem{lin2020convolution}
Zhi-Hao Lin, Sheng-Yu Huang, and Yu-Chiang~Frank Wang.
\newblock Convolution in the cloud: Learning deformable kernels in 3d graph convolution networks for point cloud analysis.
\newblock In {\em Proceedings of the IEEE/CVF conference on computer vision and pattern recognition}, pages 1800--1809, 2020.

\bibitem{vaswani2017attention}
A~Vaswani.
\newblock Attention is all you need.
\newblock {\em Advances in Neural Information Processing Systems}, 2017.

\bibitem{zhou2020cylinder3d}
Hui Zhou, Xinge Zhu, Xiao Song, Yuexin Ma, Zhe Wang, Hongsheng Li, and Dahua Lin.
\newblock {Cylinder3D}: An effective 3d framework for driving-scene lidar semantic segmentation.
\newblock {\em arXiv preprint arXiv:2008.01550}, 2020.

\bibitem{yan2018second}
Yan Yan, Yuxing Mao, and Bo~Li.
\newblock {SECOND}: Sparsely embedded convolutional detection.
\newblock {\em Sensors}, 18(10):3337, 2018.

\bibitem{garg2021seqnet}
Sourav Garg and Michael Milford.
\newblock {SeqNet}: Learning descriptors for sequence-based hierarchical place recognition.
\newblock {\em IEEE Robotics and Automation Letters}, 6(3):4305--4312, 2021.

\bibitem{caesar2020nuscenes}
Holger Caesar, Varun Bankiti, Alex~H Lang, Sourabh Vora, Venice~Erin Liong, Qiang Xu, Anush Krishnan, Yu~Pan, Giancarlo Baldan, and Oscar Beijbom.
\newblock nuscenes: A multimodal dataset for autonomous driving.
\newblock In {\em Proceedings of the IEEE/CVF conference on computer vision and pattern recognition}, pages 11621--11631, 2020.

\bibitem{geiger2013vision}
Andreas Geiger, Philip Lenz, Christoph Stiller, and Raquel Urtasun.
\newblock Vision meets robotics: The kitti dataset.
\newblock {\em The International Journal of Robotics Research}, 32(11):1231--1237, 2013.

\bibitem{liao2022kitti}
Yiyi Liao, Jun Xie, and Andreas Geiger.
\newblock Kitti-360: A novel dataset and benchmarks for urban scene understanding in 2d and 3d.
\newblock {\em IEEE Transactions on Pattern Analysis and Machine Intelligence}, 45(3):3292--3310, 2022.

\bibitem{xu2023ring++}
Xuecheng Xu, Sha Lu, Jun Wu, Haojian Lu, Qiuguo Zhu, Yiyi Liao, Rong Xiong, and Yue Wang.
\newblock Ring++: Roto-translation-invariant gram for global localization on a sparse scan map.
\newblock {\em IEEE Transactions on Robotics}, 2023.

\bibitem{lang2019pointpillars}
Alex~H Lang, Sourabh Vora, Holger Caesar, Lubing Zhou, Jiong Yang, and Oscar Beijbom.
\newblock {PointPillars}: Fast encoders for object detection from point clouds.
\newblock In {\em Proceedings of the IEEE/CVF conference on computer vision and pattern recognition}, pages 12697--12705, 2019.

\end{thebibliography}
